\documentclass[letterpaper, 10 pt, journal, twoside]{IEEEtran}

\usepackage{graphics} 
\usepackage{epsfig} 
\usepackage[ruled]{algorithm2e}
\usepackage{times} 
\usepackage{amsmath} 
\usepackage{amssymb}  

\usepackage{multirow}

\usepackage{color}

\usepackage{dblfloatfix}

\DeclareMathOperator*{\arctan2}{arctan2}

\begin{document}





\title{	BVMatch: Lidar-Based Place Recognition Using Bird's-Eye View Images}


\author{Lun Luo, Si-Yuan Cao, Bin Han, Hui-Liang~Shen, and Junwei Li
	\thanks{Manuscript received: March 17, 2021; Revised May 25, 2021; Accepted June 18, 2021. This paper was recommended for publication by Associate Editor S. Behnke and
		Editor A. Okamura upon evaluation of the Associate Editor and Reviewers' comments.
		This work was supported by the Aviation Science Foundation under Grant \#20181976. \emph{(Corresponding author: Hui-Liang Shen.)}} 
	\thanks{Lun Luo, Si-Yuan Cao, and Bin Han are with the College of Information Science and Electronic Engineering, Zhejiang University, Hangzhou 310027, China (e-mail:luolun@zju.edu.cn; karlcao@hotmail.com; binhan\_0902@foxmail.com).}%
	\thanks{Hui-Liang Shen and Junwei Li are with the College of Information Science and Electronic Engineering, Zhejiang University, Hangzhou 310027, China, and also with the Ningbo Research Institute, Zhejiang University, Ningbo 315100, China (e-mail: shenhl@zju.edu.cn; lijunwei7788@zju.edu.cn).}%
	\thanks{Digital Object Identifier (DOI): see top of this page.}
}

\markboth{IEEE Robotics and Automation Letters. Preprint Version. Accepted June, 2021}
{Luo \MakeLowercase{\textit{et al.}}: BVMatch: Lidar-Based Place Recognition Using Bird's-Eye View Images} 

\maketitle


\begin{figure*}[htbp]
	\begin{center} 
		\includegraphics [scale=0.2]{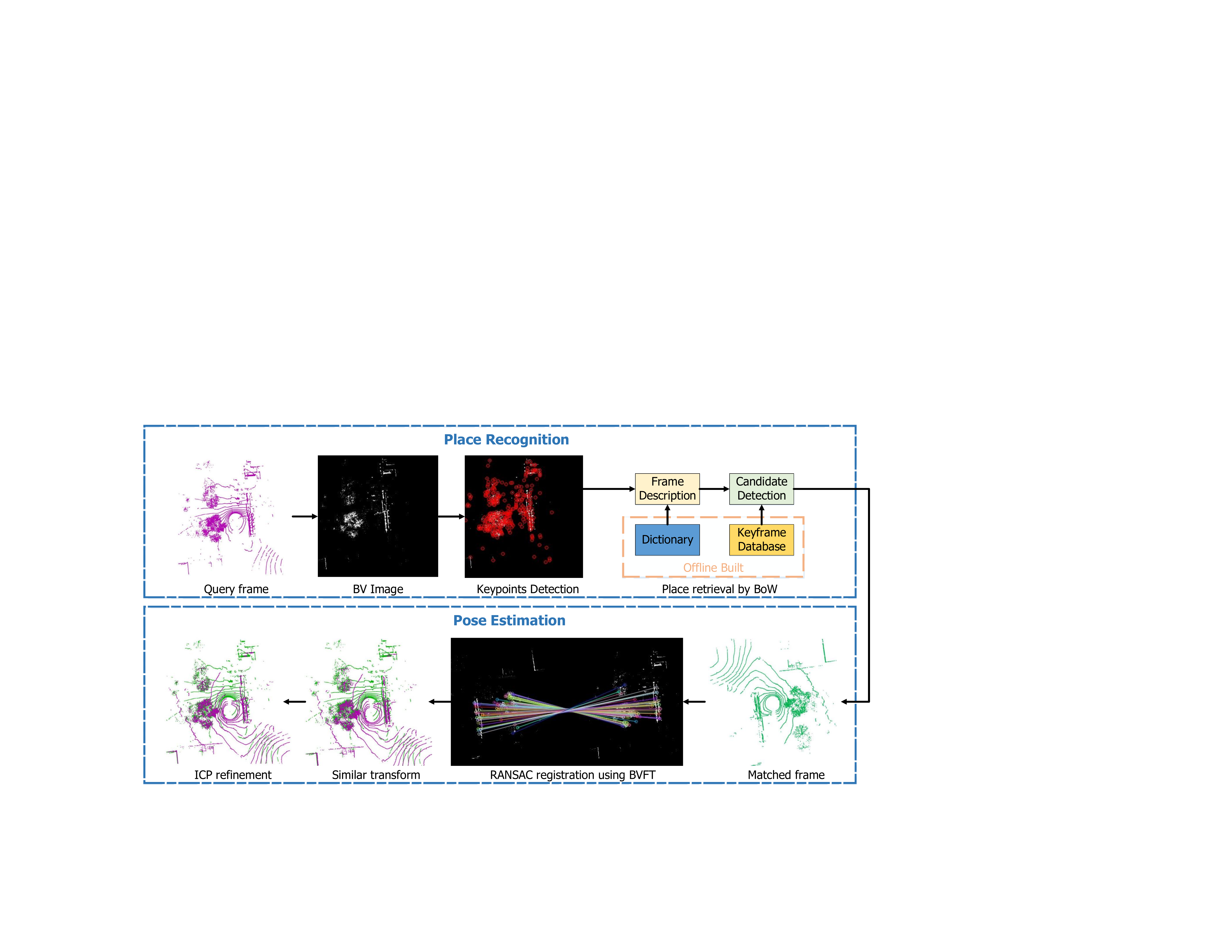}  
		\caption{BVMatch framework. In the place recognition step, the dictionary and keyframe database are built offline. Given a query frame, BVMatch generate the corresponding BV image and then extract keypoints and BVFT descriptors. Place recognition is achieved by the bag-of-words approach. In the pose estimation step, BVMatch uses RANSAC to match the BV image pair and reconstructs the 2D relative pose of the Lidar pair. Finally, BVMatch uses the 2D pose as initial guess of iterative closest point (ICP) to align the Lidar pair accurately. } 
		\label{fig:pipeline}
		
	\end{center} 
\end{figure*}

\begin{abstract}
Recognizing places using Lidar in large-scale environments is challenging due to the sparse nature of point cloud data. {In this paper we present BVMatch, a Lidar-based frame-to-frame place recognition framework, that is capable of estimating 2D relative poses. Based on the assumption that the ground area can be approximated as a plane, we uniformly discretize the ground area into grids and project 3D Lidar scans to bird's-eye view (BV) images. We further use a bank of Log-Gabor filters to build a maximum index map (MIM) that encodes the orientation information of the structures in the images. We analyze the orientation characteristics of MIM theoretically and introduce a novel descriptor called bird's-eye view feature transform (BVFT). The proposed BVFT is insensitive to rotation and intensity variations of BV images. Leveraging the BVFT descriptors, we unify the Lidar place recognition and pose estimation tasks into the BVMatch framework.} The experiments conducted on three large-scale datasets show that BVMatch outperforms the state-of-the-art methods in terms of both recall rate of place recognition and pose estimation accuracy. The source code of our method is publicly available at \texttt{https://github.com/zjuluolun/BVMatch}.
\end{abstract}


\section{Introduction}
\IEEEPARstart{P}{lace} recognition is a crucial ability for the auto-navigating robots to perform long-term {simultaneous} localization and mapping (SLAM) tasks \cite{cadena2016past}. Benefiting from the local image features \cite{2004Distinctive,rublee2011orb} and the bag-of-features techniques \cite{nister2006scalable,arandjelovic2016netvlad}, a number of image-based place recognition methods \cite{2012Bags,2017ORB} have been introduced and perform satisfactorily in some environments. However, their performance may degrade in case of illumination change and viewpoint variation, due to the nature of imaging mechanism. Compared with images, Lidar scans have larger perception fields and are insensitive to illumination changes. In road scenes where the ground area can be approximately regarded as a plane, Lidar scans are also insensitive to orientations. These advantages mean that the Lidar-based place recognition methods may have better performance than the image-based ones. Actually, current Lidar-based methods \cite{angelina2018pointnetvlad,liu2019lpd,zhang2019pcan,du2020dh3d} have shown the superiority of  Lidar scans in terms of place retrieval ability in large-scale environments. 

In this paper we present BVMatch, a novel Lidar-based place recognition method that is capable of estimating 2D relative poses. Based on the assumption that the ground area can be regarded as a plane in road scenes, BVMatch divides the ground plane into uniform grids and accumulates the scanned points in each grid to form the BV image. This orthogonal projection procedure makes the transform of the BV image pair a rigid transform, which is similar to the 2D transform of the Lidar scan pair. However, we note that BV image inherits the sparsity of Lidar scans and suffers severe intensity distortion that cannot be handled by classic image features such as SIFT \cite{2004Distinctive}. {To tackle this problem, we introduce the bird's-eye view feature transform (BVFT) that is invariant to intensity and rotation changes of BV images. We build BVFT based upon the maximum index map (MIM) of Log-Gabor filter responses \cite{li2019rift}, and theoretically analyze its characteristics on orientation shift.} Based on BVFT, BVMatch uses the bag-of-words (BoW) approach for place recognition and uses RANSAC \cite{fischler1981random} for relative pose estimation. In summary, the main contributions of this work are:
{
\begin{itemize}
\item We propose a novel local descriptor called BVFT that is insensitive to intensity and rotation variation of BV images, with which the BVMatch framework can unify the Lidar place recognition and pose estimation tasks.
\item We theoretically prove that BVFT can achieve rotation invariance  by shifting the orientations in the local MIM.
\item We experimentally validate on three large-scale datasets that BVMatch outperforms the state-of-the-arts on both place recognition and pose estimation.
\end{itemize}
}

\section{Related Work}
The Lidar-based place recognition methods can be approximately classified into two categories, i.e. the methods that directly utilize raw point clouds and the methods that use images as intermediate representations.

The first category focuses on extracting local features or global descriptors from raw point clouds. Rusu \emph{et al.} \cite{rusu2009fast} present fast point feature histograms (FPFH) to align point cloud pairs. Tombari \emph{et al.} \cite{tombari2010unique} introduce signatures of histograms for local surface description (SHOT) for surface matching. Bosse and Zlot \cite{bosse2013place} extract Gestalt keypoints and descriptors from point clouds and use keypoints voting to recognize places. {These local features exploit local characteristics of point cloud using geometric measures such as normals and curvatures, while BVFT encodes the structure information in BV images.} Instead of using local descriptors, SegMatch \cite{2018segmap} extracts higher-level segments belonging to partial or full objects, and then matches these segments for place recognition using the learned descriptors. OneShot \cite{ratz2020oneshot} follows SegMatch but extracts efficient segments from a single Lidar frame. {These two segment-based methods adopt the frame-to-map matching framework while our BVMatch is a frame-to-frame method.} M2DP \cite{he2016m2dp} projects a point cloud to multiple 2D planes and generates a density signature for points in each of the planes. The singular value decomposition (SVD) components of the signature are then used to compute a global descriptor. PointNetVLAD \cite{angelina2018pointnetvlad} leverages PointNet \cite{qi2017pointnet} to extract features of point clouds and uses NetVLAD \cite{arandjelovic2016netvlad} to generate global descriptors. {Zhang and Xiao \cite{zhang2019pcan} introduce the point contextual attention network (PCAN) that enables the network to pay more attention to the task-relevent features when generating global descriptors.} LPD-Net \cite{liu2019lpd} adopts an adaptive local feature extraction module to extract the local structures and uses a graph-based neighborhood aggregation module to discover the spatial distribution of local features. {These learning based methods do not use local keypoints and thus they cannot estimate relative poses. DH3D \cite{du2020dh3d} uses the 3D local feature encoder and detector to extract local descriptors. It embeds the descriptors to a global feature for place recognition and align the matched Lidar pairs using RANSAC. In comparison, our BVFT is handcrafted and does not need training.}

The second category projects Lidar scans to images for place recognition. Steder et al. \cite{2011Place} extract local features from range images of Lidar scans. {Unlike BV image, the range image is not Euclidean in nature since it is generated with the polar projection.} Kim \emph{et al.} \cite{kim2018scan} partition the ground space into bins according to both azimuthal and radial directions. Then they propose the scan context descriptor in a 2D matrix form. They further extend scan context to three channels, and introduce the concept of scan context image (SCI) \cite{kim20191}. Place recognition is achieved by classifying the SCIs using a convolutional network. {However, the pose estimation problem remains unsolved.} Cao \emph{et al.} \cite{cao2020season} project point clouds to cylindrical images and leverage Gabor filters to describe the contours of scenes. A histogram based descriptor is generated for every Lidar scan. {Similar to the image formations used in the aforementioned methods, the content of cylindrical image is complex due to the special projection procedure.} Thus it cannot estimate relative poses as well. To solve the problem, OverlapNet \cite{chen2020overlapnet} adopts a siamese network to estimate an overlap of range images and provides a relative yaw angle estimate of Lidar pair. However, it is incapable of estimating translation. On the contrary, BVMatch uses a simple orthogonal projection procedure that makes the transforms of BV image pair and Lidar scan pair similar, with which BVMatch can localize query Lidar scans and produce 2D poses.

{We note that BV image is similar to the occupancy grid map \cite{1989using} since they both partition space into grids. There are several methods \cite{blanco2007a,rapp2015a} designed for the occupancy grid map matching in the literature. Our BVMatch differs from these methods in two aspects. First, they target on two different types of representation. More specifically, the BV image is built with point cloud density while the occupancy grid map is formed by occupancy probabilities. Second, BVMatch is a frame-to-frame method  leveraging single 3D Lidar scans while \cite{blanco2007a,rapp2015a} are map-to-map based methods.}

\section{BV Image Based Place Recognition} 
The BVMatch method consists of the place recognition step and the pose estimation step as illustrated in Fig. \ref{fig:pipeline}. In the place recognition step, BVMatch generates the BV image and performs frame retrieval with the BoW approach. In the pose estimation step,  BVMatch matches the BV image pair using RANSAC and reconstructs the coarse 2D pose of the Lidar pair with a similar transform. At last, BVMatch uses iterative closest point (ICP) refinement with the 2D pose as an initial guess to accurately align the Lidar pair.  

 In the following, we first introduce the BV image generation mechanism and then present the coarse 2D pose reconstruction of Lidar scan pairs. Finally, we demonstrate the offline dictionary and the keyframe database creation.

\subsection{BV Image Generation}
There are several types of BV images, such as the maximum height map used in place recognition \cite{kim2018scan} and the density map used in object detection \cite{chen2017multi}. Both maps summarize the vertical shape of surrounding structures. The maximum height map uses the coordinate values of points with the maximal height, while the density map leverages the point cloud density. The maximum height map is sensitive to poses because the coordinate values of points can vary severely when robot moves. In contrast, the density map is more robust because the density of point cloud does not depend on specific points. For this reason, we use the latter as our BV image representation in this work.

Let $\mathcal{P}=\{P_i|i=1,...,N_p\}$ be a point cloud formed by points $P_i=(x_i,y_i,z_i)$ and $N_p$ the number of points in the cloud. Suppose that the point cloud is collected in road scenes. {The x-axis is pointing to the right, the y-axis is pointing forward, and the z-axis is pointing upward.} In this coordinate system, the x-y plane is the ground plane. Given a point cloud $\mathcal{P}$, we first use a voxel grid filter with the leaf size of $g$ meters to evenly distribute the points. Then we discretize the ground space into grids with resolution of $g$ meters. The point cloud density is the number of points located in each grid. We consider a [$-C $ m $, C $ m] cubic window centered at the coordinate origin. Then BV image $B(u,v)$ is a matrix of size $\lceil{\frac{2C}{g}} \rceil \times \lceil{\frac{2C}{g}}\rceil$. The BV intensity $B(u,v)$ is defined as
\begin{equation}
        \label{eq:BV image}
        \begin{aligned}
        B(u,v)=\frac{\min(N_g,N_m)}{N_m},
        \end{aligned}
\end{equation}
where $N_g$ denotes the number of points in the grid at position $(u,v)$ and $N_m$ the normalization factor. $N_m$ is set to be the 99th percentile of the point cloud density. 

BV image is a compressed representation of point cloud and describes the 2.5D structural information of an egocentric environment. It ignores the point distribution along the z-axis while keeping the rigid structures on the x-y plane. We find that the poles, facades, and guideposts in road scenes usually form edges in the BV image. These features have good repeatability and remain stable when the robot moves. We adopt FAST \cite{rosten2006machine} for feature detection and use the keypoints for registration in the BVMatch pipeline.

\subsection{Pose Reconstruction}
As BV image discretizes the ground space uniformly, the transform of the Lidar scan pair $(\mathcal{P}_i, \mathcal{P}_j)$ is similar to that of the BV image pair $(B_i(u,v),B_j(u,v))$. After obtaining the transform of $(B_i(u,v),B_j(u,v))$ we have $B_i(u,v)=B_j(u',v')$, with the coordinate
\begin{equation}
        \begin{aligned}
                &u'=\cos(\theta)u+\sin(\theta)v+t_u \\
                &v'=-\sin(\theta)u+\cos(\theta)v+t_v,
        \end{aligned}
\end{equation} 
where $(t_u,t_v,\theta)$ are transform parameters. The transform matrix $T_{ij}$ of the pair ($\mathcal{P}_i, \mathcal{P}_j)$ is 
\begin{equation}
        \label{recover}
        \begin{aligned}
                T_{ij}=  
                \begin{pmatrix} 
                        \cos(\theta) & \sin(\theta) & g{t_u} \\
                        -\sin(\theta) & \cos(\theta) & {g}{t_v} \\
                        0 & 0 & 1
                \end{pmatrix}
        \end{aligned}
\end{equation}
where $g$ denotes the leaf size of the voxel grid filter.

\begin{figure*}[htbp]
	\begin{center} 
		\includegraphics [width=5.8 in]{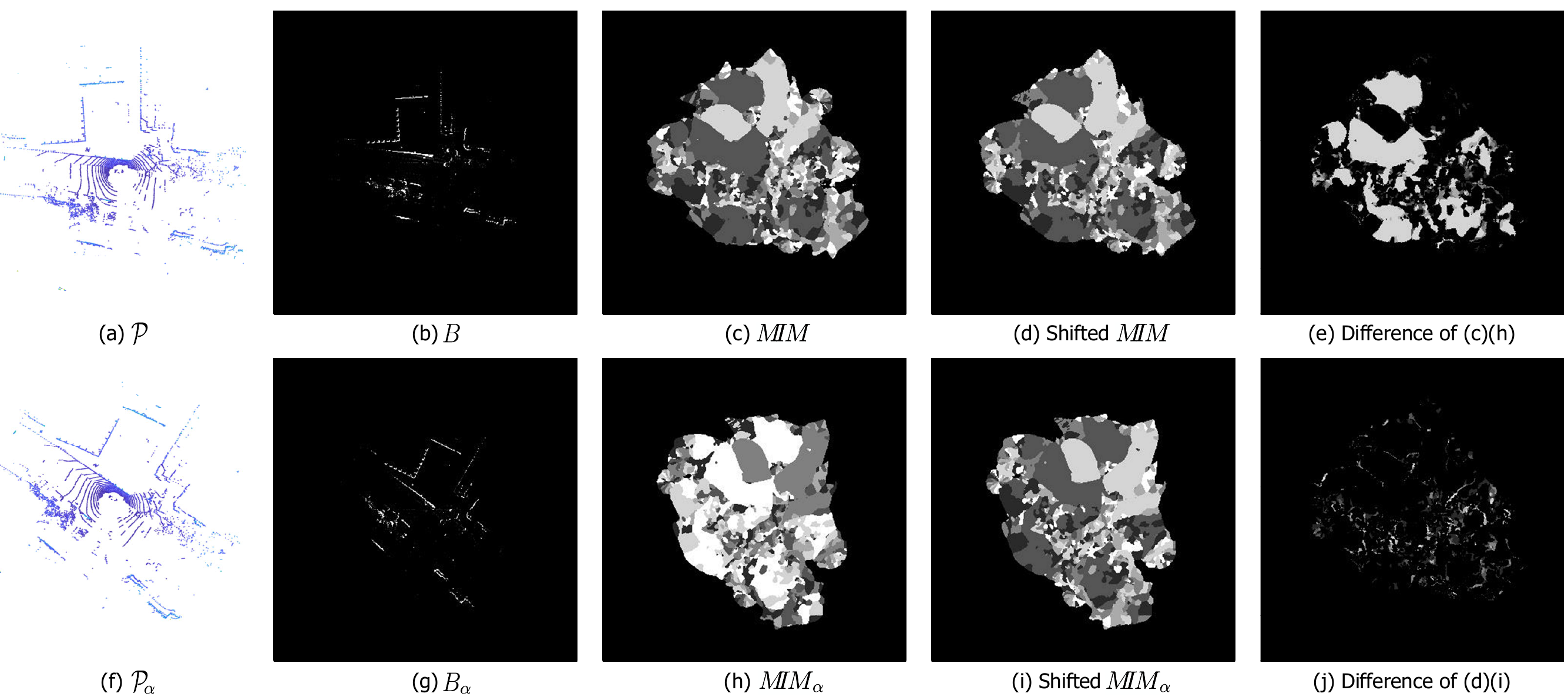}  
		\caption{ A simple example illustrating the MIM rotation invariance. $\mathcal{P}_\alpha$ is the rotated point cloud of $\mathcal{P}$. $B$ and $MIM$ are the BV image and MIM of $\mathcal{P}$, respectively. $B_\alpha$ and $MIM_\alpha$ are the BV image and MIM of $\mathcal{P}_\alpha$, respectively. It can be observed that the content of $MIM$ and rotated $MIM_\alpha$ are quite different. By shifting (i), the difference map (e) tends to be zero which confirms the rotation invariance design.} 
		\label{fig:mim}
	\end{center} 
\end{figure*}

\subsection{Dictionary and Keyframe Database}

BVMatch leverages the bag-of-words approach to extract global descriptors and uses a keyframe database to detect the best match frame of a query Lidar scan.

Bag-of-words approaches assume that similar structures in the environment produce similar distributions of features. To learn this distribution in our scene, we extract a rich set of BVFT descriptors from the training BV images. We then apply K-means clustering to obtain a total of $b$ clusters. Each cluster is a word, and the centroids of these words form a bag-of-words dictionary. We use the dictionary to encode a point cloud in terms of words by classifying the BVFT descriptors of the point cloud which words they are. To weight those words that are frequent and less discriminative, we use frequency-inverse document frequency (TF-IDF) \cite{nister2006scalable}. Finally, we get a global descriptor with size $b$ for every point cloud.

Keyfame database stores BV images with their global poses and descriptors. A robot traverses a specific place and collects Lidar scans along the way. By building a map of the place using SLAM or GPS information, every Lidar scan collected in this traversal is tagged with a global pose. We extract keyframe Lidar scans every $S$ meters the robot moves and generate a global descriptor for every keyframe. The keyframe database is built using all these global descriptors, poses, and BV images.

\section{Proposed BVFT Descriptor}

Although BV image preserves the vertical structures that are stable in a scene, it suffers severe intensity distortion due to the sparsity nature of Lidar scans. To extract distinct local descriptors, we first leverage Log-Gabor filters to compute the local responses of BV images. We then construct a maximum index map (MIM) \cite{li2019rift} originally used for multi-modal image matching. Finally, we build bird's-eye view feature transform (BVFT) that is insensitive to intensity and rotation variations of BV images. 

\subsection{Maximum Index Map (MIM)}
For simplicity,  we use the polar coordinate to represent images in the following. The polar coordinate  $(\rho,\theta)$  of a Euclidean coordinate $(u,v)$ is defined as 
\begin{equation}
        \label{eq:polar coordinate}
        \begin{aligned}
        &\rho=\sqrt{(u-\bar u)^2+(v-\bar v)^2} \\
        &\theta = \arctan2(v-\bar v,u-\bar u),\\
        \end{aligned}
\end{equation}
where $(\bar u,\bar v)$ denotes the center position of an image.

We build MIM based on Log-Gabor filters \cite{fischer2007self}. The 2D Log-Gabor filter in frequency domain is defined as 
\begin{equation}
        \label{eq: Log-Gabor Frequency}
        \begin{aligned}
\mathcal{L}(f,\omega,s,o) =\exp\left({{\frac {-(\log(f/f_s))^{2}}{2(\log(\sigma_f/f_s))^{2}}}}\right) \exp\left({\frac{{-(\omega -\omega _{o})^{2}}}{2\sigma _{\omega }^{2}}}\right),
        \end{aligned}
\end{equation}
where $f_s$ and $\omega_o$ are the center frequencies at scale $s$ and orientation $o$, $\sigma_f$ and $\sigma_\omega$ are width parameters. For a Log-Gabor filter set of $N_s$ scales and $N_o$ orientations, the filters are designed to evenly cover the spectrum. In our implementation, we follow the scales setting in \cite{li2019rift}. We select the orientations from the set $\mathcal{O}=\{0, \pi/N_o,2\pi/N_o,...,(N_o-1)\pi/N_o\}$ and hence $\omega _{o}=o\pi/N_o$. Note that the frequency part of the 2D log-Gabor filter is isotropic, which is essential for BVFT design.

We use a bank of Log-Gabor filters of $N_s$ scales and $N_o$ orientations to build the MIM. For a Log-Gabor filter at scale $s$ and orientation $o$,   Let  $L(\rho,\theta,s,o)$ be the corresponding filter of (\ref{eq: Log-Gabor Frequency}) in the spatial domain, $A(\rho,\theta,s,o)$ be the Log-Gabor amplitude response of $B(\rho,\theta)$ at orientation $o$ and scale $s$, we have
\begin{equation}
        \label{eq: magnitude response}
        A(\rho,\theta,s,o)=||B(\rho,\theta)* L(\rho,\theta,s,o)||_2,
\end{equation}
where $*$ represents the convolution operation. Then the Log-Gabor amplitude response at orientation $o$ is
\begin{equation}
        \label{eq: orientation response}
A (\rho,\theta,o)=\sum_s A(\rho,\theta,s,o).
\end{equation}
The maximum index map (MIM) is a map of the orientations with the maximal Log-Gabor responses,  that is
\begin{equation}
        \label{eq:MIM}
MIM (\rho,\theta)=\arg\max_o A (\rho,\theta,o).
\end{equation}

\subsection{BVFT Descriptor} 
The Log-Gabor filters, widely employed in image processing, are jointly localized in position, orientation and spatial frequency \cite{fischer2007self}. In BVFT, we employ Log-Gabor filters to capture the orientation information of the rigid vertical structures in scenes using (\ref{eq:MIM}). This makes MIM insensitive to intensity distortion of BV images. However, MIM is still sensitive to rotation. In the following we show that, by shifting the orientations in the MIM with respect to a specified dominant orientation, rotation invariance can be achieved.

Fig. \ref{fig:mim}(a) is a randomly selected Lidar scan $\mathcal{P}$ from the Oxford RobotCar Radar dataset \cite{barnes2020oxford}. Fig. \ref{fig:mim}(f) is the rotated point cloud $\mathcal{P}_\alpha$ of $\mathcal{P}$ with a rotation angle $\alpha$ about the z-axis. Fig. \ref{fig:mim}(g) and (b) are the corresponding BV images $B_\alpha$ and $B$ related as 
\begin{equation}
        \label{eq: BV relation}
        B_\alpha(\rho,\theta) = B(\rho,\theta+\alpha).
\end{equation}
Fig. \ref{fig:mim}(h) and (c) are $MIM_\alpha$ and $MIM$ respectively. It can be observed that the contents of $B$ and $B_\alpha$ are the same, while the contents of $MIM$ and $MIM_\alpha$ are quite different. To eliminate the rotation difference, we rotate $MIM_\alpha$ by $-\alpha$ and obtain Fig. \ref{fig:mim}(i). It is observed the difference of two MIMs shown in Fig. \ref{fig:mim}(d) is significant. To obtain rotation invariance, this difference should be eliminated.

We compute the Log-Gabor amplitude response of $B_\alpha$ with orientation $o$, and have 
\begin{equation}
        \label{eq: orientation response relation 1}
        \begin{aligned} 
        A_{\alpha}(\rho,\theta,o) 
        &=\sum_s ||B_\alpha(\rho,\theta)* L(\rho,\theta,s,o)||_2 \\
        & =\sum_s ||B(\rho,\theta+\alpha)* L(\rho,\theta,s,o)||_2.
        \end{aligned}
\end{equation}
Since the frequency part of the 2D log-Gabor filter is isotropic in the frequency domain (see Eq. (\ref{eq: Log-Gabor Frequency})), every filter at scales $s$ can be obtained by rotating other filter at the same scale by some angle. When $\alpha$ is an angle within the orientation set $\mathcal{O}$,
\begin{equation}
        \label{eq: sp LGF relation freqeuncy}
        \mathcal{L}(f,\omega,s,o) = \mathcal{L}(f,\omega+\alpha,s,o_\alpha),
\end{equation}
where 
{\begin{equation}
o_\alpha=\mathrm{mod}(o-\alpha\frac{N_o}{\pi},N_o). 
\end{equation}}
Here we use a $\mathrm{mod}$ operation on the orientation because the $N_o$ orientations form a ring structure \cite{li2019rift}. Since 2D Fourier transform is rotation invariant, the 2D spatial Log-Gabor filter has the same property as the frequency one, 
\begin{equation}
        \label{eq: sp LGF relation}
        L(\rho,\theta,s,o) = L(\rho,\theta+\alpha,s,o_\alpha).
\end{equation}
Substituting (\ref{eq: sp LGF relation}) to (\ref{eq: orientation response relation 1}) yields
\begin{equation}
        \label{eq: orientation response relation 2}
        \begin{aligned} 
        A_{\alpha}(\rho,\theta,o) 
        &=\sum_s ||B(\rho,\theta+\alpha)*L(\rho,\theta+\alpha,s,o_\alpha)||_2\\
        &=A(\rho,\theta+\alpha,o_\alpha).
        \end{aligned}
\end{equation}
This means that the Log-Gabor response image $A_{\alpha}(\rho,\theta,o)$ can be obtained by rotating $A(\rho,\theta+\alpha,o_\alpha)$ by $\alpha$. Substituting (\ref{eq: orientation response relation 2}) to  (\ref{eq:MIM}) yields
\begin{equation}
        \label{eq: MIM relation}
        \begin{aligned}
        MIM_\alpha(\rho,\theta)
        &=\arg\max_o A_\alpha(\rho,\theta,o)  \\
        &=\arg\max_o A(\rho,\theta+\alpha,o_\alpha) \\ 
        &=\mathrm{mod} (MIM(\rho,\theta+\alpha)-\alpha\frac{N_o}{\pi}, N_o),
        \end{aligned}
\end{equation}
which reveals that when the point cloud is rotated by any angle within the orientation set $\mathcal{O}$, its MIM can be obtained by a simple circle shift operation on the MIM of the unrotated one. Fig. \ref{fig:mim} validates the above analysis. Fig. \ref{fig:mim}(j) is the rotated and shifted $MIM_\alpha$ using Eq. (\ref{eq: MIM relation}). It can be observed from Fig. \ref{fig:mim}(e) that the difference map is very small, in despite of the quantization error of orientation set $\mathcal{O}$.


We further extend the above analysis onto image patches and design rotation invariant local descriptors. For every detected keypoint, we find its dominant orientation. We build a local histogram $h(o)$ of pixel values over a square MIM patch $\mathrm{patch}(\rho,\theta)$ with $J\times J$ pixels centered at the keypoint. When computing the histogram, the increments are weighted by a Gaussian window function centered at the keypoint with mean $(0,0)$ and standard deviation $(J/2,J/2)$. Suppose that the peak of the histogram is at orientation $o_{m}$, i.e., $o_{m}=\arg\max\limits_{o} h(o)$, the dominant orientation is
\begin{equation}
        \label{eq: dominant orientation}
\beta = \pi\frac{o_{m}}{N_o}.
\end{equation}
Then we rotate the patch by $\beta$ and shift the patch,
\begin{equation}
        \label{eq: patch relation}
        \mathrm{patch}_\beta(\rho,\theta)=\mathrm{mod}(\mathrm{patch}(\rho,\theta+\beta)-o_{m},N_o).
\end{equation}
With this treatment, the local patch is rotation invariant. We divide the patch into $l\times l$ sub-grids and build a distribution histogram for each sub-grid. These histograms are concatenated to form a BVFT feature vector of size $l\times l\times N_o$.

The above analysis is conducted under the assumption that the rotation angles are within the orientation set $\mathcal{O}$. However, (\ref{eq: sp LGF relation freqeuncy}) and (\ref{eq: MIM relation}) still hold when the rotation angles are within the set $\mathcal{\hat{O}}=\{\pi, (N_o+1)\pi/N_o,(N_o+2)\pi/N_o,...,(2N_o-1)\pi/N_o\}$. Thus, we cannot determine whether the dominant orientation is $\pi\frac{o_{m}}{N_o}$ or $(2\pi-\pi\frac{o_{m}}{N_o})$ given $o_{m}$. To avoid this ambiguity, we rotate every shifted MIM patch by $\pi$ and assign every keypoint with an additional descriptor generated using this patch. When the rotation angle is not within $\mathcal{O}$ or $\mathcal{\hat{O}}$, it is obvious that (\ref{eq: MIM relation}) does not hold. In fact, BVFT can not cover continuous orientation because the Log-Gabors are band-pass filters. In practice, we have found that BVFT has good matching ability in continuous orientation, and we will show that our BVFT has excellent performance for the place recognition problem in the experiment section.

\begin{table*}\scriptsize
	\renewcommand\arraystretch{1.2}
	\renewcommand\tabcolsep{5pt}
	\begin{center}
		\caption{Training and test sets of the datasets}
		\label{table:dataset_partition}
		\begin{tabular}{|c|c|c|}
			\hline
			\hline
			  \multicolumn{3}{|c|}{Dataset: NCLT}                             \\ \hline
			 & Train Set               & Test Set                             \\ \hline
			Seq.    & \begin{tabular}[c]{@{}c@{}}2013-02-23,2013-04-05\\ \end{tabular}                    & \begin{tabular}[l]{@{}l@{}}2012-01-15, 2012-02-04, 2012-03-17, 2012-05-26, 2012-06-15, 2012-08-20, 2012-09-28, 2012-10-28, 2012-11-16 \end{tabular}    \\ \hline

			Frames    & \begin{tabular}[c]{@{}c@{}} 2050\quad\qquad 2036  \end{tabular}                    & \begin{tabular}[l]{@{}l@{}} 1402  \qquad 1050 \qquad\quad 1062 \qquad\quad  1155\qquad\quad\  711\qquad\quad\ \   1103\qquad\quad  1012 \qquad\quad  1040 \qquad\quad 873 \end{tabular}    \\ \hline \hline
					\multicolumn{3}{|c|}{Dataset: Oxford Radar RobotCar}                             \\ \hline	
			 & Train Set               & Test Set                             \\ \hline
			Seq.   & \begin{tabular}[c]{@{}c@{}}2019-01-10-11-46,2019-01-11-12-26\end{tabular} & \begin{tabular}[l]{@{}l@{}} 2019-01-14-12-05-52, 2019-01-15-13-53-14, 2019-01-16-14-15-33, 2019-01-17-14-03-00, 2019-01-18-12-42-34  \end{tabular}  
			 \\ \hline
			 Frames   & \begin{tabular}[c]{@{}c@{}} 2334 \quad\qquad 2133 \end{tabular} & \begin{tabular}[l]{@{}l@{}} 
			 	 1181\qquad\qquad  	\qquad\quad\quad 1164 	\qquad\qquad\qquad\ \  1132 \qquad\qquad\qquad\ \  1222 \qquad\qquad\qquad\ \ \ \  1164 \end{tabular}  
			 \\ \hline
			 \hline
		\end{tabular}
	\end{center}
\end{table*}

\begin{figure*}
	\centering
	\includegraphics[width=6in]{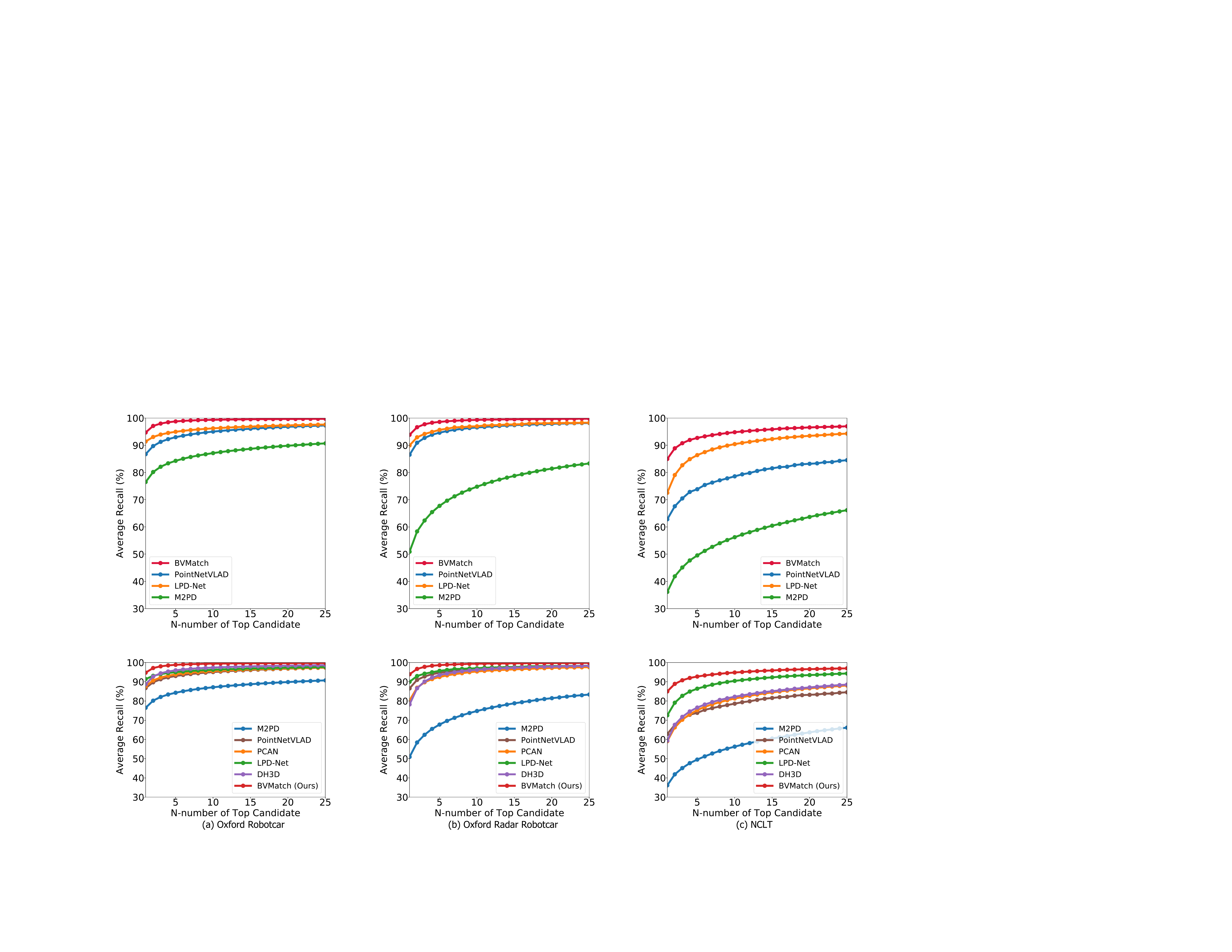}
	\caption{Average recalls with respect to the N-number of top candidates on different datasets.}
	\label{fig: topN} 
\end{figure*}


\begin{table}[!htb]
	\renewcommand\arraystretch{1.2}
	\renewcommand\tabcolsep{5pt}
	\newcommand{\tabincell}[2]{\begin{tabular}{@{}#1@{}}#2\end{tabular}}
	\begin{center}
		\caption{Average recalls at Top-1\% and Top-1}
		\label{table: top1}
		\begin{tabular}{|c|c|c|c|c|c|c|c|}
			\hline\hline
			& \multicolumn{2}{c|}{Oxford} & \multicolumn{2}{c|}{Oxford Radar} & \multicolumn{2}{c|}{NCLT} \\ \hline
			& @1\%        & @1       & @1\%           & @1       & @1\%           & @1     \\ \hline
			M2DP \cite{he2016m2dp}    &85.8   & 76.5                     &  76.4  &50.9               & 57.5          &    36.2                    \\ \hline
			PN-VLAD \cite{angelina2018pointnetvlad} &  96.8   & 86.7                     & 94.7            & 86.6          &    78.3     & 62.8                      \\ \hline
			PCAN \cite{zhang2019pcan} &  97.0   & 87.7                     & 93.8            & 80.3          &    82.2     & 59.0                      \\ \hline
			LPD-Net \cite{liu2019lpd} & 97.4      & 91.4                   & 95.5     &  90.0                 & 90.7     & 72.5                           \\ \hline
			DH3D \cite{du2020dh3d} &  97.8   & 88.7                     & 94.6            & 78.2          &    83.0     & 59.4                      \\ \hline
			BVMatch (Ours) & \textbf{98.8}      & \textbf{95.7}           & \textbf{99.2}      & \textbf{93.9}        & \textbf{95.2}         & \textbf{83.6}        \\ \hline
			\hline
		\end{tabular}
	\end{center}
\end{table}

\section{Experiments}
We compare the place recognition capability of our BVMatch with M2DP \cite{he2016m2dp}, PointNetVLAD \cite{angelina2018pointnetvlad}, PCAN \cite{zhang2019pcan}, LPD-Net \cite{liu2019lpd} and DH3D \cite{du2020dh3d}, among which the last four methods are deep learning ones. We compare the pose estimation performance with SIFT \cite{2004Distinctive} (e.g., extracting SIFT descriptors from BV images and using RANSAC for pose estimation), OverlapNet \cite{chen2020overlapnet} and DH3D. Note that DH3D can perform both place recognition and pose estimation like BVMatch. The source codes of all the competing methods are publicly available on the websites.\footnote{https://github.com/LiHeUA/M2DP\\https://github.com/mikacuy/pointnetvlad\\https://github.com/XLechter/PCAN\\https://github.com/Suoivy/LPD-net\\https://github.com/JuanDuGit/DH3D\\https://github.com/Suoivy/OverlapNet} For the sake of fairness, we fine-tuned all the deep learning methods on the training sets of the individual datasets. For M2DP and SIFT, we use the parameters provided by the authors. For BVMatch, we empirically set the parameters as follows: voxel grid size $g=0.4$ meters,  Log-Gabor filter frequency scales $N_s=4$ and orientations $N_o=6$, local patch size $J=96$, number of subgrids $l\times l=6\times 6$, and size of bag-of-words dictionary $b=10000$. Accordingly, the size of BVFT descriptor is $l\times l\times N_o=216$. 

\subsection{Datasets}
We conduct the experiments on three long-term and large-scale datasets: Oxford RobotCar dataset \cite{barnes2020oxford}, Oxford RobotCar Radar dataset \cite{maddern20171} and NCLT dataset \cite{2016University}. 

\emph{1) Oxford RobotCar dataset:} The Oxford RobotCar dataset was created using a 2D LMS LiDAR sensor mounted on a car that repeatedly drives through Oxford at different times traversing a 10 km route. It captures many different combinations of weather, traffic, and pedestrians over one year. However, it only provides 2D Lidar scans. To make a 3D point cloud with enough points, we use the relative pose between a previous scan and a recent scan to accumulate sequential 2D Lidar scans with a trajectory length of 80 m. The relative pose is obtained from GPS/IMU readings. We select 45 sequences from the dataset for evaluating (the same sequences used in PointNetVLAD \cite{angelina2018pointnetvlad}). 

\emph{2) Oxford RobotCar Radar dataset:} The Oxford RobotCar Radar dataset was created at the same place as the Oxford RobotCar dataset in 7 days. It provides sparse 3D Lidar point clouds generated by two Velodyne32-VLP LiDAR sensors mounted on the left and right sides of a car. In this work, we only use the data of the left Lidar. Since the sequences collected on the same day are quite similar, we randomly select a sequence from each day and get 7 sequences.

\emph{3) NCLT dataset:} The NCLT dataset was created at the University of Michigan North Campus using a Velodyne32-HDL LiDAR sensor with varying routes. It provides sparse 3D Lidar point clouds. We use 11 sequences from the dataset to evaluate the methods.

Note that the three datasets are of different characteristics. Specifically, the Lidar scans in the Oxford Radar RobotCar dataset and the NCLT dataset are sparser than those in the Oxford RobotCar dataset due to the usage of different types of Lidar sensor. We split the sequences in every dataset into training and testing sequences. For the Oxford RobotCar dataset, we use 15 sequences collected in 2014 for training and 30 sequences collected in 2015 for test. For the Oxford RobotCar Radar dataset and the NCLT dataset, the sequence partition is summarized in Table \ref{table:dataset_partition}. All the training sequences are sampled every 2 meters, while the test sequences are sampled every 10 meters. Point cloud data in a [$-50 $ m, $ 50 $ m] cubic window of a frame are used for all methods.

\begin{figure}
    \centering
    \includegraphics[width=3.1in]{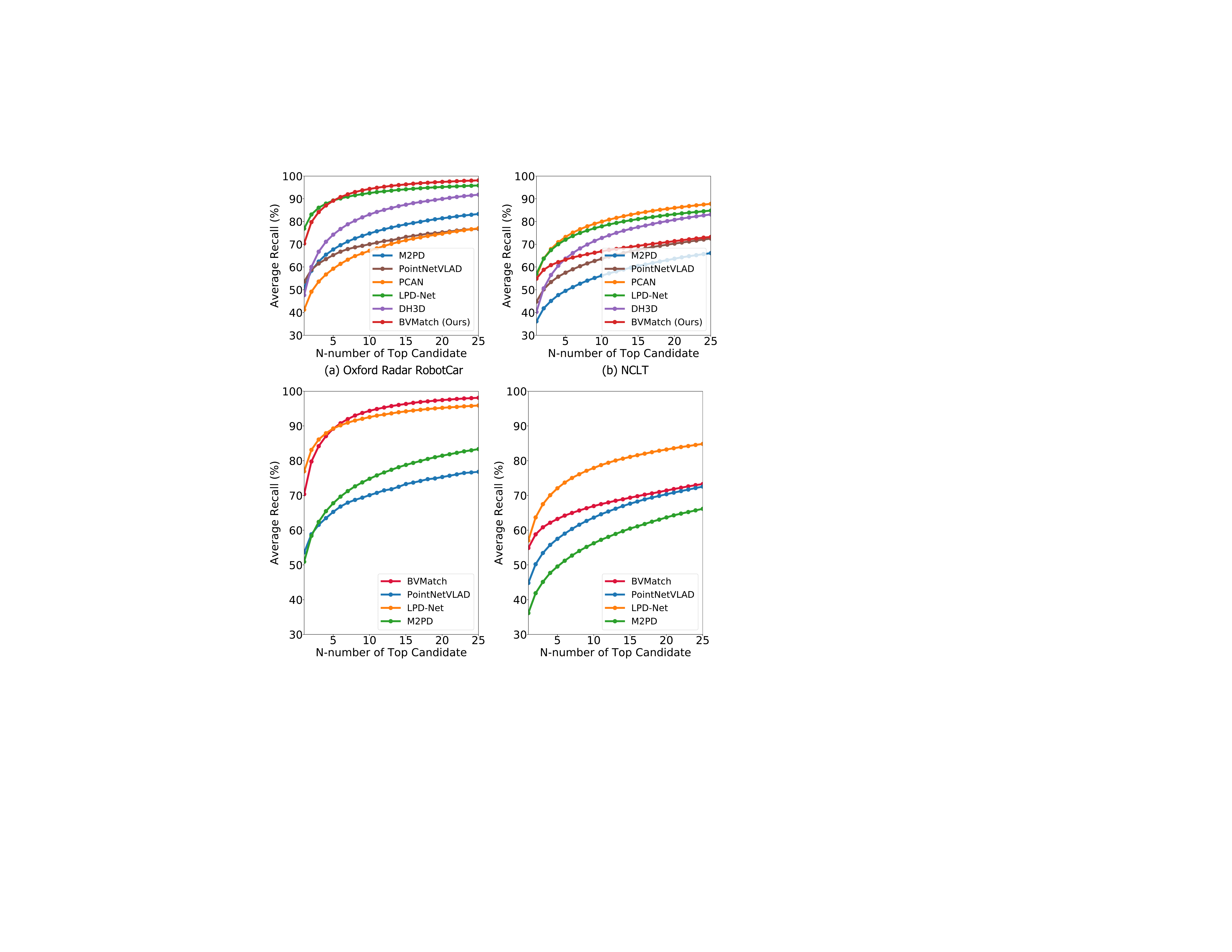}
    \caption{Evaluation of generalization ability in terms of average recalls with respect to the N-number of top candidates. The methods are trained on the Oxford RobotCar dataset and tested on the Oxford Radar Robotcar and NCLT datasets.}
    \label{fig:generalization_ability} 
\end{figure}

\begin{table}[!htb]
	\renewcommand\arraystretch{1.2}
	\renewcommand\tabcolsep{5pt}
	\newcommand{\tabincell}[2]{\begin{tabular}{@{}#1@{}}#2\end{tabular}}
	\begin{center}
		\caption{Average recalls at Top-1\% and Top-1 on the Oxford Radar RobotCar and NCLT datasets for generalization ability evaluation}
		\label{table: generalization}
		\begin{tabular}{|c|c|c|c|c|}
			\hline\hline
			& \multicolumn{2}{c|}{Oxford Radar} & \multicolumn{2}{c|}{NCLT} \\ \hline
			& @1\%        &  @1       &  @1\%           &  @1          \\ \hline
			M2DP \cite{he2016m2dp}    & 76.4         & 50.9    & 57.5         & 36.2      \\ \hline
			PN-VLAD \cite{angelina2018pointnetvlad} & 71.8                      & 53.6                   & 64.2                   & 44.8                \\ \hline
			PCAN \cite{zhang2019pcan} & 69.0  & 41.3  & \textbf{81.2}      & 55.9                \\ \hline
			LPD-Net \cite{liu2019lpd} & 93.2    & \textbf{76.9}   & {78.4}      & \textbf{57.1}    \\ \hline
			DH3D \cite{du2020dh3d} & 85   & 47.7      & 74.3      & 40.3      \\ \hline
			BVMatch (Ours) & \textbf{95.3}             & {70.3}          & {67.8}          & {54.9}       \\ 
			\hline \hline 
		\end{tabular}
	\end{center}
\end{table}

\begin{table*}[!hbp]
	\renewcommand\arraystretch{1.2}
	\renewcommand\tabcolsep{5pt}
	\newcommand{\tabincell}[2]{\begin{tabular}{@{}#1@{}}#2\end{tabular}}
	\begin{center}
		\caption{Pose estimation Results  on the Oxford robotcar and  the NCLT dataset}
		\label{table: pose estimation}
		\begin{tabular}{|c|c|c|c|c|c|c|c|c|c|c|}
			\hline\hline
			& \multicolumn{5}{c|}{Oxford RobotCar} & \multicolumn{5}{c|}{NCLT} \\ \hline
			& \multicolumn{2}{c|}{RTE (m)}        &  \multicolumn{2}{c|}{RRE (deg)}       &  SR (\%)          &  \multicolumn{2}{c|}{RTE (m)} & \multicolumn{2}{c|}{RRE (deg)} & SR (\%)          \\ \hline
			&Mean&Std. & Mean&Std. &  & Mean&Std. & Mean&Std. &    \\ \hline
			SIFT \cite{2004Distinctive}   & {0.77}&{0.47} & 0.53 &0.47 & {82.1} & 0.69&0.46 & 1.49&1.23 &53.3    \\ \hline
			OverlapNet \cite{chen2020overlapnet}   & \multicolumn{5}{c|}{-}& \multicolumn{2}{c|}{-} & 2.43&1.42  & 15.4        \\ \hline
			DH3D \cite{du2020dh3d}   & \textbf{0.43}&\textbf{0.31} & 1.23&0.89 & \textbf{99.9} & 1.16&0.54 & 3.43&1.06 &14.8    \\ \hline
			BVMatch (Ours) & 0.51&0.33 & \textbf{0.41}&\textbf{0.42}  & 99.1 &  \textbf{0.57}&\textbf{0.38} & \textbf{1.08}&\textbf{1.00}  & \textbf{94.5}    \\ 
			\hline\hline
		\end{tabular}
	\end{center}
\end{table*}
\subsection{Place Recognition}

We first train the bag-of-words dictionary in each dataset with the training sequences. We then build keyframe databases with global descriptors for the test sequences. For every keyframe in a sequence, we use the Euclidean distance of the global descriptors to retrieve the best match from other sequences. The match is positive if the ground truth distance of the matched frame is less than $t$ meters. We use Top-N recall to evaluate the place recognition ability. Fig. \ref{fig: topN} shows the average recall results at Top-N with the ground truth threshold distance $t=25$ meters. Table \ref{table: top1} illustrates the average recalls at Top-1 and 1\%. It can be observed that our BVMatch outperforms other methods in every dataset.

To validate the generalization ability of the methods, we train the deep learning based methods and the bag-of-words dictionary on the Oxford Robotcar dataset and evaluate the methods on the other two datasets. {Fig. \ref{fig:generalization_ability} illustrates that our BVMatch outperforms the competitors on the Oxford Radar dataset while is of moderate performance on the NCLT dataset}. {Table \ref{table: generalization} further shows that BVMatch ranks first at Top-1\% and ranks second at Top-1 on the Oxford Radar dataset, but does not perform very well on the NCLT dataset.}

\begin{figure}
	\centering
	\includegraphics[width=3.1in]{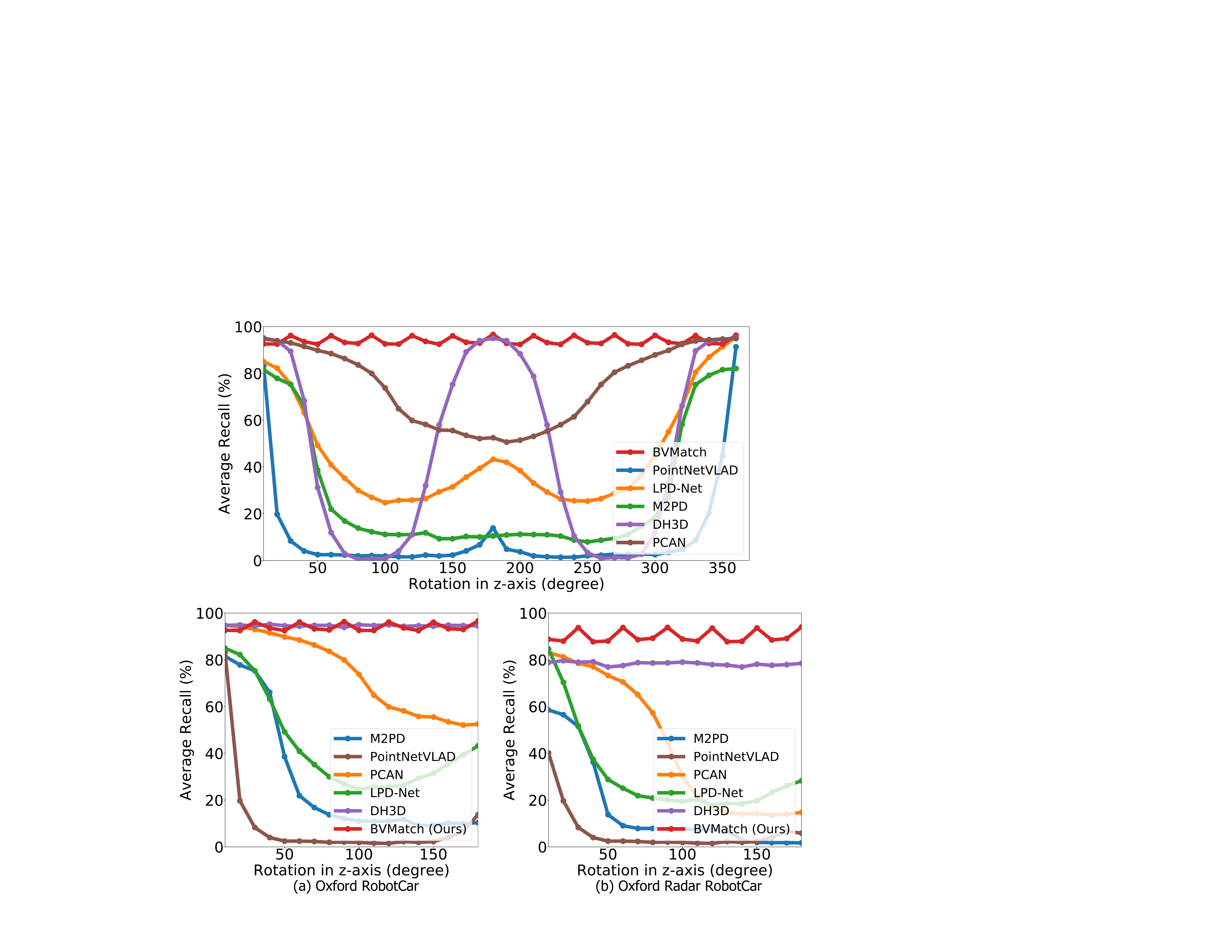}
	\caption{Average recalls with respect to varying orientations on the two Oxford datasets. }
	\label{fig: rotation invariance} 
\end{figure}

\begin{figure}
	\centering
	\includegraphics[width=3.1in]{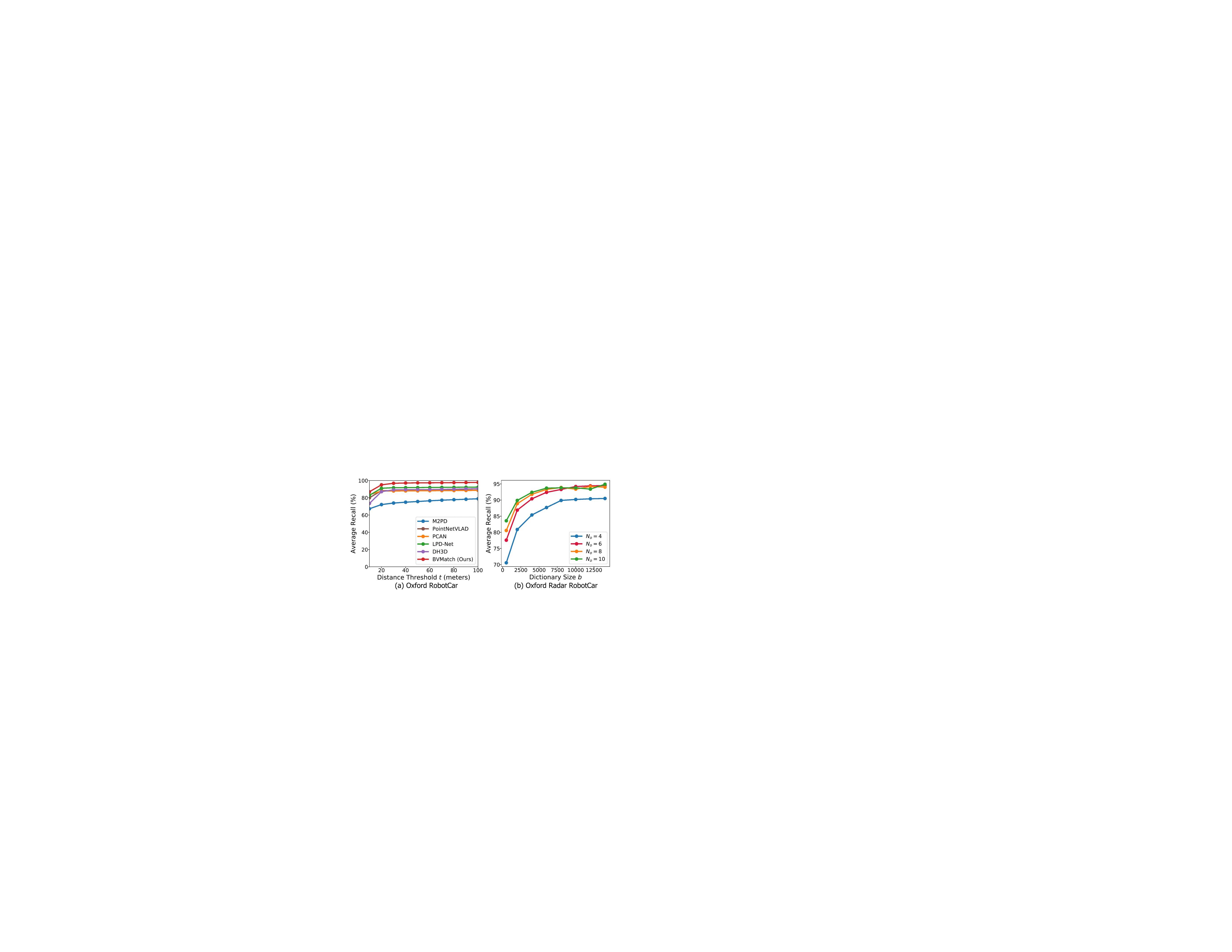}
	\caption{Parameter study. (a) Average recalls at Top-1  with respect to varying distance thresholds on the Oxford RobotCar dataset. (b) Average recalls at Top-1 with respect to orientation numbers $N_o$ and dictionary size $b$ on the the Oxford Radar RobotCar dataset.}
	\label{fig:parameter_study} 
\end{figure}

Robustness to orientation variance is important since the robot may have arbitrary headings in real-world applications. Since BVFT is rotation invariant, BVMatch is inherently orientation invariant. We evaluate place retrieval ability of the methods with varying orientations on the sequences \texttt{2015-02-03} and \texttt{2015-02-10} in the Oxford RobotCar dataset and {on the sequences \texttt{2019-01-14-12-05-52} and \texttt{2019-01-15-13-53-14} in the Oxford Radar RobotCar dataset}. The average recall at Top-1 at every rotation angle is shown in Fig. \ref{fig: rotation invariance}. It is observed that the recall rates of DH3D are relatively stable while other deep learning methods drop dramatically when the angle deviates from $0^\circ$. Our BVMatch performs closely to DH3D on the Oxford RobotCar dataset, but much better than DH3D on the Oxford Radar RobotCar dataset in which the Lidar scans are more sparse. Note that the recall curve of BVMatch is fluctuant due to the angle quantization of the orientation set $\mathcal{O}$. 

An extension evaluation on retrieval ability with respect to the distance threshold $t$ is shown in Fig. \ref{fig:parameter_study}(a). It is clear that BVMatch performs better than other methods under various distance threshold settings.

We conducted an experiment on the Oxford Radar RobotCar dataset to investigate the influence of Log-Gabor filter orientations $N_o$ and the dictionary size $b$. Fig. \ref{fig:parameter_study}(b) shows that when $N_o$ and $b$ increase, the average recall at Top-1 grows and saturate at about $N_o=6$ and $b=10000$.



\subsection{Pose Estimation}

%
%
%
We evaluate the performance of pose estimation on the sequences \texttt{2015-02-03} and \texttt{2015-02-10} of the Oxford RobotCar dataset and on the sequences \texttt{2012-02-04} and \texttt{2012-01-15} of the NCLT dataset. In the evaluation, we retrieve the best match of each query frame in a sequence (e.g., \texttt{2015-02-03}) from the other sequence in the same dataset (e.g., \texttt{2015-02-10}). We obtain 752 matched pairs from the Oxford dataset and 969 pairs from the NCLT dataset. {Since the ground truth poses in the Oxford dataset are biased, we align the Lidar pairs by ICP and regard the aligned results as the ground truth.

Our BVMatch uses the BVFT descriptor with RANSAC to find the relative poses. We compared BVMatch with SIFT \cite{2004Distinctive}, OverlapNet \cite{chen2020overlapnet} and DH3D \cite{du2020dh3d}. Since OverlapNet is only suitable for single 3D scans and cannot use the data accumulated from 2D Lidar scans in the Oxford dataset, we only evaluate OverlapNet on the NCLT dataset. For fair comparison, we fine-tuned OverlapNet and DH3D on the datasets. We compute the relative translation error (RTE) and the relative rotation error (RRE). We regard the pose as a successful estimation when the RTE and RRE are below 2 meters and $5^\circ$, respectively. Table \ref{table: pose estimation} shows that, on the Oxford dataset, our BVMatch achieves the best RRE while performs slightly worse than DH3D on the RTE and SR metrics. On the NCLT dataset, BVMatch outperforms the competitive methods by a large margin. This further validates the superiority of BVMatch in handling sparse Lidar scans.

\subsection{Runtime Evaluation}
We implemented the BVFT descriptor generation and RANSAC using C++ and implemented BVMatch framework using Python. Our platform running the experiments is a desktop computer equipped with an Intel Quad-Core 3.40 GHz i5-7500 CPU and 16 GB RAM. The average time cost for each frame is 0.29 seconds to extract BVFT descriptors, 0.02 seconds to generate a global descriptor, 0.05 seconds to perform retrieval, and 0.23 seconds to register BV image pair using RANSAC. The total time cost for each frame is 0.59 seconds. 

\section{Conclusions}

This paper has presented BVMatch, a novel Lidar-based frame-to-frame place recognition method. The method employs bird's-eye view (BV) image as the intermediate representation of point cloud, and introduces the BVFT descriptor to perform match. Compared with the state-of-the-arts, BVMatch is more efficient in place recognition and is able to estimate 2D relative poses. In our future work, we will focus on further reducing its time complexity and improving its generalization ability.

\bibliography{ref}
\bibliographystyle{ieeetr}

\end{document}